\documentclass{article}

\usepackage{arxiv}

\usepackage[utf8]{inputenc} % allow utf-8 input
\usepackage[T1]{fontenc}    % use 8-bit T1 fonts
\usepackage{hyperref}       % hyperlinks
\usepackage{url}            % simple URL typesetting
\usepackage{booktabs}       % professional-quality tables
\usepackage{amsfonts}       % blackboard math symbols
\usepackage{nicefrac}       % compact symbols for 1/2, etc.
\usepackage{microtype}      % microtypography
\usepackage{lipsum}		% Can be removed after putting your text content
\usepackage{graphicx}
\usepackage{natbib}
\usepackage{doi}
\usepackage{amsmath}
\usepackage{tikz}
\usepackage{geometry}
\usepackage{tikz}
\usetikzlibrary{positioning, arrows.meta, shapes.misc}
\usetikzlibrary{mindmap, shadows, positioning, backgrounds}
\usetikzlibrary{positioning, shapes, arrows.meta}
\usepackage{amsmath, amssymb}
\title{The SAM2-to-SAM3 Gap in the Segment Anything Model Family: Why Prompt-Based Expertise Fails in Concept-Driven Image Segmentation}

\author{%
\href{https://orcid.org/0000-0002-5417-6744}{\includegraphics[scale=0.06]{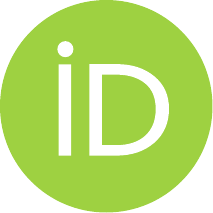}\hspace{1mm}Ranjan Sapkota\textsuperscript{1*}} \quad
\href{https://orcid.org/0000-0002-8098-1616}{\includegraphics[scale=0.06]{orcid.pdf}\hspace{1mm}Konstantinos I.~Roumeliotis\textsuperscript{2}} \quad
\href{https://orcid.org/0000-0001-5337-4848}{\includegraphics[scale=0.06]{orcid.pdf}\hspace{1mm}Manoj Karkee\textsuperscript{1}} \\
\textsuperscript{1}Cornell University, Ithaca, NY 14850, USA \\
\textsuperscript{2}University of the Peloponnese, Tripoli 22131, Greece \\
}

\renewcommand{\shorttitle}{SAM2-to-SAM3 Gap in the Segment Anything Model}

\hypersetup{
pdftitle={A template for the arxiv style},
pdfsubject={q-bio.NC, q-bio.QM},
pdfauthor={David S.~Hippocampus, Elias D.~Striatum},
pdfkeywords={First keyword, Second keyword, More},
}

\begin{document}
\maketitle

\begin{abstract}
This paper investigates the fundamental discontinuity between the latest two Segment Anything Models: SAM2 and SAM3. We explain why the expertise in prompt-based segmentation of SAM2 does not transfer to the multimodal concept-driven paradigm of SAM3. SAM2 operates through spatial prompts points, boxes, and masks yielding purely geometric and temporal segmentation. In contrast, SAM3 introduces a unified vision-language architecture capable of open-vocabulary reasoning, semantic grounding, contrastive alignment, and exemplar-based concept understanding. We structure this analysis through five core components: (1) a Conceptual Break Between Prompt-Based and Concept-Based Segmentation, contrasting spatial prompt semantics of SAM2 with multimodal fusion and text-conditioned mask generation of SAM3; (2) Architectural Divergence, detailing pure vision-temporal design of SAM2 versus integration of vision-language encoders, geometry and exemplar encoders, fusion modules, DETR-style decoders, object queries, and ambiguity-handling via Mixture-of-Experts in SAM3; (3) Dataset and Annotation Differences, contrasting SA-V video masks with multimodal concept-annotated corpora of SAM3; (4) Training and Hyperparameter Distinctions, showing why SAM2 optimization knowledge does not apply to SAM3; and (5) Evaluation, Metrics, and Failure Modes, outlining the transition from geometric IoU metrics to semantic, open-vocabulary evaluation. Together, these analyses establish SAM3 as a new class of segmentation foundation model and chart future directions for the emerging concept-driven segmentation era. This project repository is available on GitHub (~\href{https://github.com/Applied-AI-Research-Lab/The-SAM2-to-SAM3-Gap-in-the-Segment-Anything-Model-Family}{\texttt{Link}}).
\end{abstract}

% keywords can be removed
\section{Introduction}

Image segmentation, the process of partitioning an image into semantically meaningful regions forms the backbone of modern computer vision and automation systems \cite{liu2019recent, guo2018review}. At its core, segmentation enables machines to identify, isolate, and interpret objects, surfaces, and boundaries at the pixel level \cite{ghosh2019understanding, blaschke2004image, minaee2021image}, transforming raw visual data into structured representations suitable for downstream reasoning and decision-making \cite{yang2021bottom, manakitsa2024review, archana2024deep}. 

Likewise, object segmentation, a specialized form of segmentation, goes beyond detecting the presence of objects: it delineates precise contours and regions of interest, enabling fine-grained analysis of shapes, attributes, textures, and spatial relationships \cite{gao2023deep, yao2020video}. Because segmentation provides pixel-level understanding, it is indispensable across numerous industries where accuracy, reliability, and contextual awareness are essential \cite{xu2024advances}. Applications spans many sectors: medical and biomedical imaging (tumor and organ segmentation) \cite{seo2020machine, roth2018deep, maulik2009medical, conze2023current}, autonomous driving and intelligent transportation (lane detection, vehicle and pedestrian segmentation) \cite{chen2023edge, chen2021deep, zakaria2023lane, ahmed2019pedestrian}, agriculture and environmental monitoring (plant organs, weeds, fruitlets, soil health) \cite{meenakshi2022soil, luo2024semantic, lei2024deep, anand2021agrisegnet}, manufacturing and industrial inspection (defect detection, quality control) \cite{usamentiaga2022automated, mirbod2022industrial, huang2015automated, ferguson2018detection}, robotics and automation (object grasping, scene parsing) \cite{du2021vision, uckermann20123d, ainetter2021end, colleoni2021robotic}, satellite and remote sensing (land cover mapping, infrastructure monitoring) \cite{rogan2004remote, song2018segment, ramos2024multispectral, jia2024semantic}, augmented and virtual reality (scene reconstruction) \cite{gonzalez2020nextmed, yang2013image, gonzalez2020nextmed}, and security and surveillance (object tracking and anomaly detection) \cite{abdullah2023semantic, gruosso2021human, he2020segmentations}. In each domain, segmentation acts as the interpretive layer that bridges perception and action, enabling tasks such as diagnosis, navigation, manipulation, planning, measurement, and anomaly detection \cite{manakitsa2024review}. Its importance continues to grow as visual systems shift from handcrafted pipelines toward foundation models that demand massive annotated datasets and semantic consistency \cite{awais2025foundation, chen2025weakly}. Within this landscape, advances in segmentation architectures directly influence the reliability, generalizability, and autonomy of AI-driven systems.

Over the past several decades, object segmentation has evolved through multiple generations of computer vision algorithms, beginning with classical methods such as thresholding \cite{kohler1981segmentation, kumar2018thresholding}, region growing \cite{tilton2012best, roggero2002object}, active contours (Snakes) \cite{xu2007object, chen2023overview}, graph cuts \cite{xu2007object, campbell2010automatic, boykov2006graph}, watershed segmentation \cite{levner2007classification, hall2004detecting, peng2010object, mangan1999partitioning}, and Markov Random Fields (MRFs) \cite{kato2012markov, liu2017deep, zheng2017semantic, anguelov2005discriminative}. With the rise of deep learning, convolutional neural networks (CNNs) transformed segmentation accuracy through architectures like Fully Convolutional Networks (FCN) \cite{dai2016r}, U-Net \cite{ronneberger2015u}, Mask R-CNN \cite{he2017mask, dollar2017mask}, DeepLab (v1–v3+) \cite{chen2017deeplab, liu2019auto}, PSPNet \cite{zhao2017pyramid}, and SegNet \cite{badrinarayanan2017segnet}, each introducing advances in multiscale feature extraction, contextual aggregation, and instance-level reasoning. 

More recent segmentation approaches incorporated transformer-based architectures (e.g., DETR \cite{zhu2020deformable, dai2021dynamic, li2022dn}, SegFormer \cite{xie2021segformer}, Mask2Former \cite{cheng2022masked}), advanced attention mechanisms (multi-head attention \cite{cordonnier2020multi}, cross-attention \cite{lin2022cat}, deformable attention \cite{xia2022vision}), and self-supervised learning strategies such as MAE(Masked Autoencoders \cite{he2022masked}) BYOL Bootstrap Your Own Latent \cite{grill2020bootstrap}) and SimCLR(A Simple Framework for Contrastive Learning of Visual Representations \cite{chen2020simple}). Despite these advancements, traditional segmentation models typically require task-specific training, domain adaptation, or finely tuned architectures to generalize effectively \cite{toldo2020unsupervised, wang2020differential, zoetmulder2022domain}. The Segment Anything Model (SAM) family breaks from this paradigm by offering promptable, zero-shot segmentation at foundation scale, enabling masks to be generated without retraining and across domains unseen during training. This shift marks a fundamental transition from model-centric segmentation pipelines to universal, prompt-driven segmentation frameworks.

\begin{figure}[ht]
\centering
\begin{tikzpicture}[
    node distance=5mm and 12mm,
    version/.style={rectangle, draw, rounded corners, minimum width=3.2cm, minimum height=0.9cm, align=center, font=\small\bfseries},
    task/.style={rectangle, draw, rounded corners, minimum width=3.2cm, minimum height=0.75cm, font=\footnotesize, align=center},
    unsupported/.style={task, dashed},
    arrowstyle/.style={-{Stealth}, thick}
]

% --- Version nodes (timeline) ---
\node[version] (sam1) {SAM1 \\ (2023)};
\node[version, right=3.2cm of sam1] (sam2) {SAM2 \\ (2024)};
\node[version, right=3.2cm of sam2] (sam3) {SAM3 \\ (2025)};

% --- Timeline arrows ---
\draw[arrowstyle] (sam1) -- (sam2);
\draw[arrowstyle] (sam2) -- (sam3);

% --- SAM1 capabilities (vertical stack) ---
\node[task, below=6mm of sam1] (s1img)   {Promptable Image\\Segmentation};
\node[task, below=3mm of s1img] (s1zero) {Zero-Shot Transfer\\on Images};
\node[task, below=3mm of s1zero] (s1annot) {Interactive\\Annotation Tools};
\node[unsupported, below=3mm of s1annot] (s1video) {Video Segmentation\\\& Tracking};
\node[unsupported, below=3mm of s1video] (s1concept) {Concept/Text-Based\\Segmentation};

\draw[arrowstyle] (sam1) -- (s1img);
\draw[arrowstyle] (s1img) -- (s1zero);
\draw[arrowstyle] (s1zero) -- (s1annot);
\draw[arrowstyle] (s1annot) -- (s1video);
\draw[arrowstyle] (s1video) -- (s1concept);

% --- SAM2 capabilities (vertical stack) ---
\node[task, below=6mm of sam2] (s2video) {Promptable Video\\Segmentation};
\node[task, below=3mm of s2video] (s2track) {Online Object\\Tracking};
\node[task, below=3mm of s2track] (s2mem)   {Temporal Memory\\\& Mask Propagation};
\node[unsupported, below=3mm of s2mem] (s2concept) {Open-Vocabulary\\Concept Segmentation};
\node[unsupported, below=3mm of s2concept] (s2text) {Text-Only\\Prompts};

\draw[arrowstyle] (sam2) -- (s2video);
\draw[arrowstyle] (s2video) -- (s2track);
\draw[arrowstyle] (s2track) -- (s2mem);
\draw[arrowstyle] (s2mem) -- (s2concept);
\draw[arrowstyle] (s2concept) -- (s2text);

% --- SAM3 capabilities (vertical stack) ---
\node[task, below=6mm of sam3] (s3concept) {Concept-Level\\Segmentation};
\node[task, below=3mm of s3concept] (s3open) {Open-Vocabulary\\Text Prompts};
\node[task, below=3mm of s3open] (s3fusion) {Multimodal\\Vision--Language Fusion};
\node[task, below=3mm of s3fusion] (s3examples) {Example-Based\\Concept Learning};
\node[task, below=3mm of s3examples] (s3auto) {Automatic Instance\\Discovery};

\draw[arrowstyle] (sam3) -- (s3concept);
\draw[arrowstyle] (s3concept) -- (s3open);
\draw[arrowstyle] (s3open) -- (s3fusion);
\draw[arrowstyle] (s3fusion) -- (s3examples);
\draw[arrowstyle] (s3examples) -- (s3auto);

\end{tikzpicture}
\caption{Timeline of Segment Anything models (SAM1 to SAM3) and their core capabilities. Solid boxes denote natively supported functionalities; dashed boxes denote capabilities that are unsupported or only achievable via external components.}
\label{fig:sam_timeline_capabilities_compact}
\end{figure}
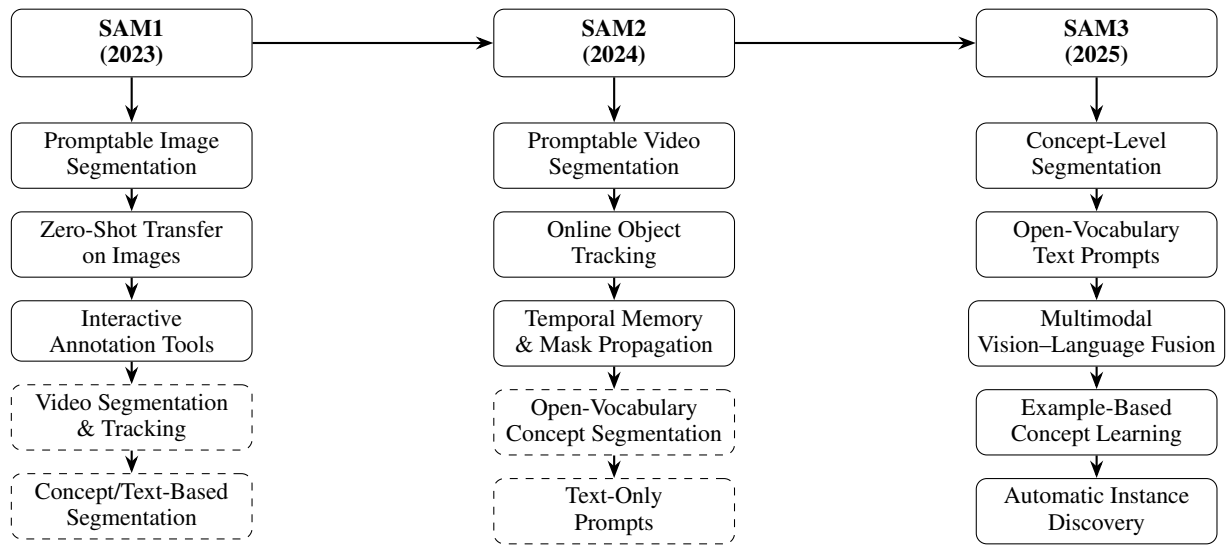

The Segment Anything family (SAM) reshaped modern segmentation by introducing prompt-driven, general-purpose vision models capable of zero-shot mask generation across diverse domains \cite{kirillov2023segment}. When SAM1 was released in 2023 as depicted in Figure \ref{fig:sam_timeline_capabilities_compact}, it established a new paradigm in which segmentation could be initiated using simple visual cues such as points, boxes, or masks \cite{kirillov2023segment}. SAM1’s innovation lay in its large ViT-based encoder and geometric prompt fusion mechanism, enabling rapid mask extraction without task-specific training. Building on this, SAM2 extended the framework into video via temporal memory structures and consistent mask propagation \cite{ravi2024sam}. However, the release of SAM3 fundamentally altered the paradigm by introducing concept-driven multimodal segmentation creating a discontinuity that renders SAM2 experience insufficient for SAM3 fine-tuning or deployment \cite{carion2025sam}. Figure \ref{fig:sam_timeline_capabilities_compact} illustrates the evolutionary timeline of SAM and their capabilities.

\section{Conceptual Break Between Prompt-Based and Concept-Based Segmentation}
\begin{figure}[h!]
     \centering
     \includegraphics[width=0.95\linewidth]{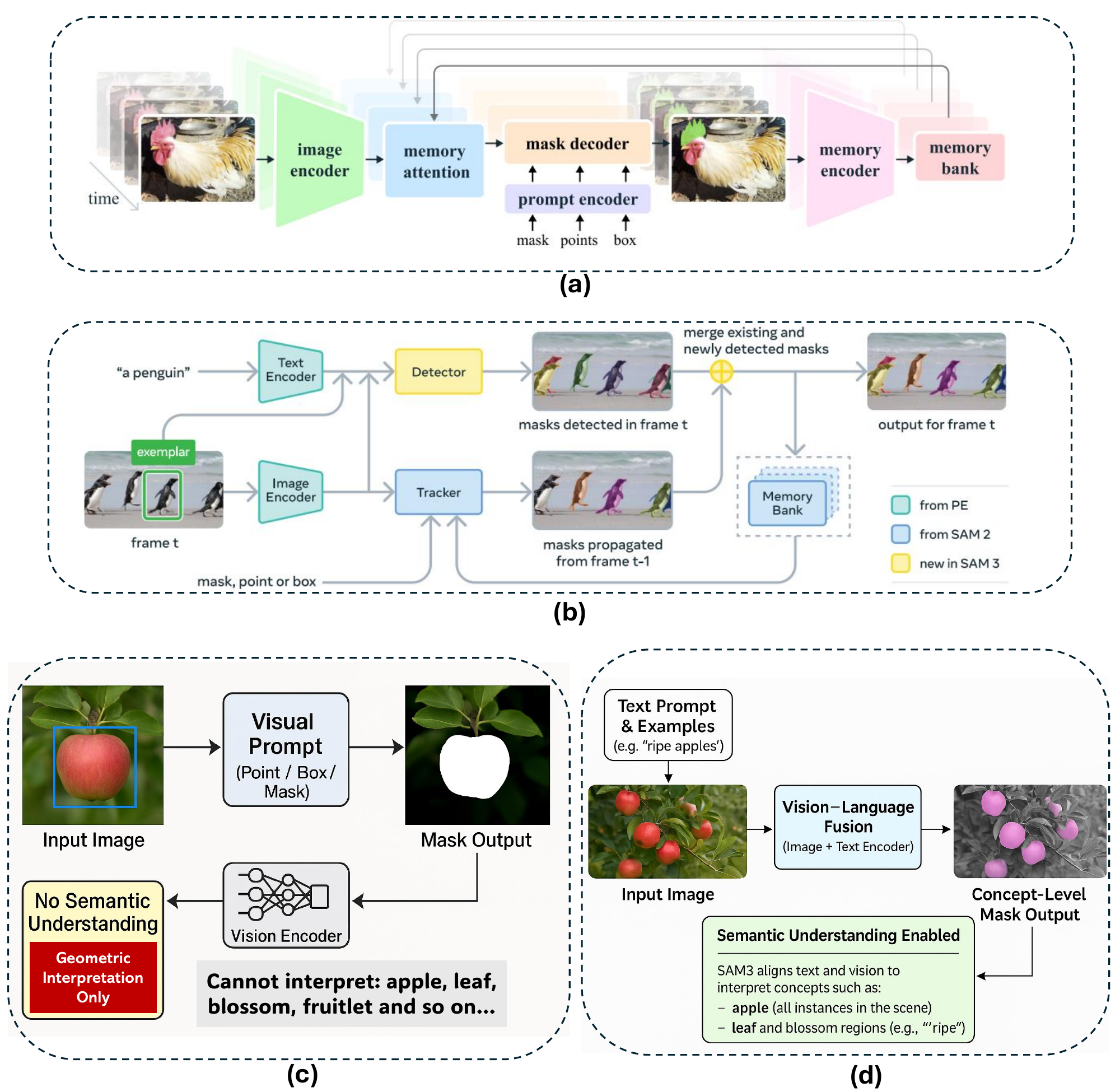}
    \caption{(a) SAM2 architecture: a prompt-driven vision–temporal pipeline where segmentation depends on spatial prompts and memory retrieval across frames ; (b) SAM3 architecture: a multimodal vision-language system with a new detector enabling open-vocabulary, concept-level segmentation  ;(c) SAM2 orchard workflow showing prompt-based apple segmentation without semantic understanding  ;  and (d) SAM3 workflow demonstrating text-prompted, concept-aware segmentation, identifying all relevant apple instances through multimodal fusion.}
    \label{fig:architectural}
\end{figure}

The transition from SAM2 to SAM3 represents one of the most substantial paradigm shifts in segmentation research since the emergence of prompt-based foundation models \cite{carion2025sam}. As illustrated in Figure~\ref{fig:architectural}a, SAM2 extends the original SAM1 framework into real-time video segmentation by incorporating a temporal memory mechanism; however, the model remains entirely dependent on spatial visual prompts. Users must still provide points, bounding boxes, or initial mask priors to specify the target region \cite{ravi2024sam}, reinforcing SAM2’s reliance on geometric guidance rather than semantic understanding. This limitation is further exemplified in Figure~\ref{fig:architectural}c, where SAM2 generates segmentation masks for orchard apples only after receiving explicit spatial cues.

In contrast, SAM3 as shown in Figure~\ref{fig:architectural}b introduces a multimodal vision-language detector capable of open-vocabulary reasoning, enabling segmentation driven by textual concepts rather than manual geometric prompts. Figure~\ref{fig:architectural}d demonstrates this advancement: SAM3 autonomously identifies and segments all concept-relevant apple instances from a natural-language prompt, illustrating a fundamental shift from prompt-based object localization to concept-driven semantic interpretation.

This transition from SAM2 to SAM3 represents one of the most substantial paradigm shifts in segmentation foundations since the introduction of prompt-based segmentation \cite{carion2025sam}. Although SAM2 expanded the original Segment Anything Model (SAM1) framework into real-time video segmentation through temporal memory, it still relied solely on visual prompting: users provided points, boxes, or mask priors to indicate what to segment \cite{ravi2024sam}. In contrast, SAM3 introduced a unified multimodal framework capable of interpreting text, concepts, and example images as segmentation instructions \cite{ravi2024sam, carion2025sam}. This shift unlocked open-vocabulary segmentation and semantic reasoning beyond explicit human clicks. Understanding this difference is crucial for any researcher training or deploying SAM3, particularly those familiar only with SAM2's prompt-driven paradigm.

\subsection{Prompt Semantics in SAM2: Spatial but Not Semantic}
SAM2 requires explicit visual grounding before segmentation \cite{zhu2024medical}. The user must indicate where an object resides through spatial prompts, typically by adding a point on the target, drawing a bounding box around the region of interest, or providing an initial mask that SAM2 refines \cite{zhou2025edgesam, zhang2025two, sun2025efficient}. The model interprets these cues purely as geometric signals. No internal semantic reasoning is performed \cite{zhao2024inspiring, yu2025crisp}; For example  in orchard images, SAM2 does not inherently know what an apple, leaf, branch, blossom, or fruitlet is. It instead refines the visible structure around the prompted region based on learned visual patterns and segmentation priors \cite{gutierrez2025prompt, wang2025optimizing}. This makes SAM2 highly effective for interactive editing, video object tracking, and manual annotation workflows, where a human or external system provides specific locations. However, the lack of semantic grounding limits its applicability in automated systems that require concept-level understanding \cite{zeng2025efficientsam3, carion2025sam}, such as orchard fruit load estimation, defect detection, or selective harvesting, where objects and attributes must be discovered without manual clicks.

Figure \ref{fig:architectural} illustrates the fundamental conceptual and architectural transition from prompt-driven segmentation in SAM2 to concept-driven multimodal reasoning in SAM3. In Figure \ref{fig:architectural}a, the SAM2 architecture is shown as a strictly vision-temporal pipeline in which segmentation predictions depend on the current visual prompt and previously stored memory embeddings. Each frame is streamed through an image encoder, cross-attended with memory features from earlier frames, and then passed to a mask decoder that refines the segmentation for that frame. Importantly, as the figure highlights, SAM2’s prompt encoder accepts only spatial cues-points, boxes, or initial masks and the model operates entirely within a geometric interpretation space. This design enables highly stable temporal mask propagation but precludes any understanding of object semantics or categories.

Figure \ref{fig:architectural}b contrasts this with SAM3’s expanded architecture, where a new detector module (highlighted in yellow in the original diagram) incorporates text embeddings produced by a language encoder. The detector introduces open-vocabulary grounding, enabling the system to identify and segment concepts (“penguin” in the illustration), while the tracker fuses example-based visual prompts with the multimodal representations. SAM3’s memory bank now stores semantically enriched tokens, allowing concept-level segmentation even when objects undergo appearance changes. This architectural shift from pure vision processing to vision-language fusion represents the core break that invalidates direct transfer of SAM2 expertise to SAM3.

Figures \ref{fig:architectural}c and \ref{fig:architectural}d provide concrete orchard-image examples that further clarify the conceptual distinction. In Figure \ref{fig:architectural}c, SAM2 segments an apple only when explicitly prompted with a bounding box, demonstrating no semantic understanding: the system cannot interpret whether the region contains an apple, leaf, blossom, or fruitlet. The output mask is simply a geometric refinement of the user’s spatial selection. In contrast, Figure \ref{fig:architectural}d shows SAM3 segmenting all ripe apples from a text prompt (“ripe apples”). Because SAM3 fuses image and text embeddings, semantic reasoning becomes available, enabling the model to differentiate between apples and surrounding foliage and to detect all concept-relevant instances automatically.

\subsection{Concept Semantics in SAM3: Multimodal Fusion and Open-Vocabulary Reasoning}
SAM3 accepts textual prompts such as ``segment all ripe apples'' and optionally accepts example images as concept demonstrations. These instructions are interpreted semantically rather than spatially, requiring the model to align visual evidence with linguistic embeddings. SAM3 learns a shared representation space in which visual features and text embeddings interact, enabling concept-level detection and segmentation. Instead of waiting for explicit human localization, SAM3 autonomously searches the image for features that match the textual description. This transforms segmentation from a spatial interaction problem into a semantic inference problem. Performance now depends on multilingual embeddings, concept alignment quality, and vision language fusion depth, rather than only on the quality of spatial prompts. As a result, expertise that is rooted in click strategies, bounding-box design, or temporal prompt propagation in SAM2 does not transfer directly when training SAM3, which operates as a concept-driven multimodal segmentation system.

\begin{figure}[ht]
\centering
\begin{tikzpicture}[
    node distance=5mm and 14mm,
    box/.style={rectangle, draw, rounded corners, minimum width=3.2cm, minimum height=0.85cm, align=center, font=\footnotesize},
    process/.style={box, fill=green!10},
    concept/.style={box, fill=blue!10},
    arrowstyle/.style={-{Stealth}, thick}
]

% --- SAM2 column ---
\node[font=\small\bfseries] (sam2title) {SAM2: Spatial Prompt Pipeline};

\node[box, below=5mm of sam2title] (s2img)  {Input Image / Video};
\node[box, below=of s2img]          (s2prompt) {Visual Prompt \\ (Point/Box/Mask)};
\node[process, below=of s2prompt]   (s2enc) {Vision Encoder};
\node[process, below=of s2enc]      (s2mem) {Temporal Memory};
\node[process, below=of s2mem]      (s2dec) {Mask Decoder};
\node[concept, below=of s2dec]      (s2out) {Prompted Object Mask};

\draw[arrowstyle] (s2img) -- (s2prompt);
\draw[arrowstyle] (s2prompt) -- (s2enc);
\draw[arrowstyle] (s2enc) -- (s2mem);
\draw[arrowstyle] (s2mem) -- (s2dec);
\draw[arrowstyle] (s2dec) -- (s2out);

% --- SAM3 column ---
\node[font=\small\bfseries, right=4.0cm of sam2title] (sam3title) {SAM3: Conceptual Multimodal Pipeline};

\node[box, below=5mm of sam3title] (s3img)  {Input Image};
\node[box, below=of s3img]         (s3text) {Text Prompt \\ (e.g., ``Ripe Apples'')};
\node[box, below=of s3text]        (s3ex)   {Example Images \\ (Optional)};
\node[process, below=of s3ex]      (s3fuse) {Vision--Language Fusion};
\node[process, below=of s3fuse]    (s3dec)  {Concept-Conditioned \\ Mask Decoder};
\node[concept, below=of s3dec]     (s3out)  {Concept-Level Masks \\ (All Matching Instances)};

\draw[arrowstyle] (s3img) -- (s3text);
\draw[arrowstyle] (s3text) -- (s3ex);
\draw[arrowstyle] (s3ex) -- (s3fuse);
\draw[arrowstyle] (s3fuse) -- (s3dec);
\draw[arrowstyle] (s3dec) -- (s3out);

\end{tikzpicture}

\caption{Compact comparison of SAM2 and SAM3 segmentation workflows. SAM2 follows a purely vision-based, spatially prompted pipeline, whereas SAM3 integrates text prompts, example images, and vision-language fusion to produce open-vocabulary, concept-level masks.}
\label{fig:sam2sam3_pipeline_compact}
\end{figure}
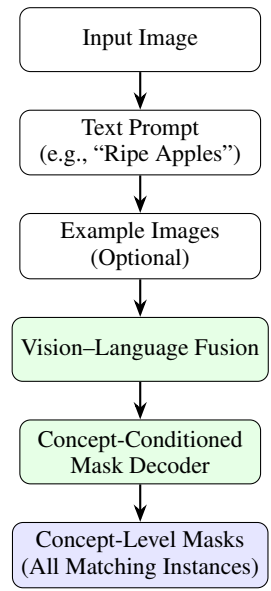

Figure \ref{fig:sam2sam3_pipeline_compact} illustrates the fundamental conceptual divergence between SAM2 and SAM3, highlighting why expertise in prompt-based spatial segmentation does not transfer naturally to concept-driven multimodal segmentation. As shown in the upper portion of Figure \ref{fig:sam2sam3_pipeline_compact}, SAM2 follows a purely vision-centric workflow in which an image or video is processed only after the user supplies explicit spatial prompts such as points, boxes, or mask priors. This pipeline emphasizes geometric localization and temporal memory, enabling SAM2 to track and refine objects selected by the user but preventing it from understanding semantic categories or attributes. In contrast, the lower branch demonstrates how SAM3 integrates text prompts, optional example images, and a unified vision-language fusion module to infer concept-level masks for all relevant instances in the scene. This multimodal design transforms segmentation from a spatial interaction task into a semantic reasoning task, enabling SAM3 to perform open-vocabulary and attribute-sensitive segmentation without manual localization. 

\section{Architectural Divergence Undermining Knowledge Transfer}
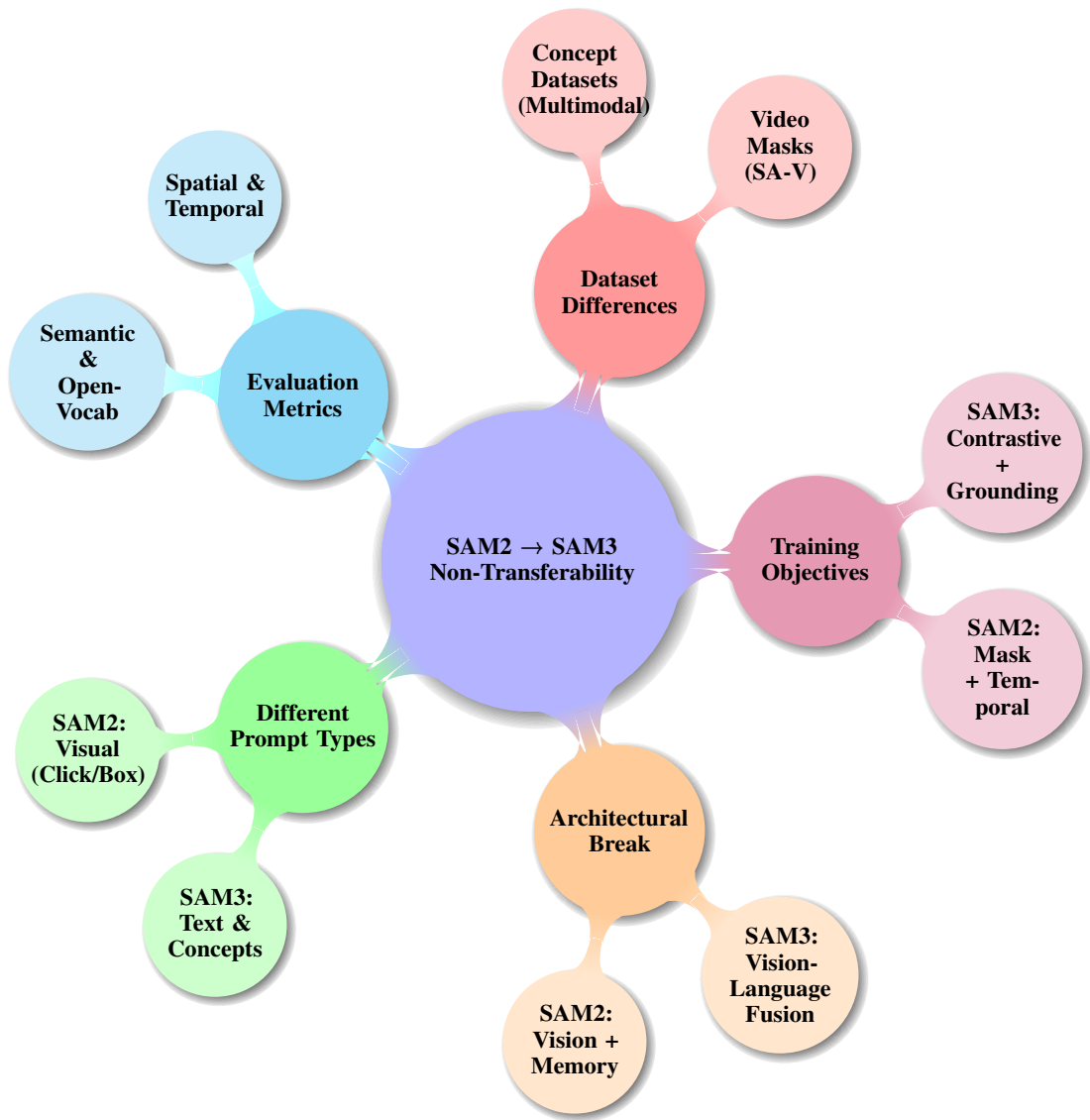
\begin{figure}[ht]
\centering
\begin{tikzpicture}[
    mindmap,
    grow cyclic,
    concept color=blue!30,
    every node/.style={concept, circular drop shadow, font=\small\bfseries, text=black},
    level 1 concept/.append style={level distance=3.8cm, sibling angle=72},
]

\node{SAM2 → SAM3 \\ Non-Transferability}
    child[concept color=green!40] {
        node {Different \\ Prompt Types}
        child[concept color=green!20]{ node {SAM2: Visual \\ (Click/Box)} }
        child[concept color=green!20]{ node {SAM3: Text \& \\ Concepts} }
    }
    child[concept color=orange!40] {
        node {Architectural \\ Break}
        child[concept color=orange!20]{ node {SAM2: Vision +  \\ Memory} }
        child[concept color=orange!20]{ node {SAM3: Vision-Language \\ Fusion} }
    }
    child[concept color=purple!40] {
        node {Training \\ Objectives}
        child[concept color=purple!20]{ node {SAM2: Mask \\ + Temporal} }
        child[concept color=purple!20]{ node {SAM3: \\ Contrastive \\ + \\Grounding} }
    }
    child[concept color=red!40] {
        node {Dataset \\ Differences}
        child[concept color=red!20]{ node {Video Masks \\ (SA-V)} }
        child[concept color=red!20]{ node {Concept Datasets \\ (Multimodal)} }
    }
    child[concept color=cyan!40] {
        node {Evaluation \\ Metrics}
        child[concept color=cyan!20]{ node {Spatial \& Temporal} }
        child[concept color=cyan!20]{ node {Semantic \& \\ Open-Vocab} }
    };

\end{tikzpicture}

\caption{Mindmap summarizing the scientific reasons why expertise in SAM2 does not transfer to SAM3. The gap arises from differences in prompting, architecture, training objectives, datasets, and evaluation metrics.}
\label{fig:sam2sam3_mindmap}
\end{figure}

Figure \ref{fig:sam2sam3_mindmap} provides a unified conceptual overview of why knowledge and training intuition gained from SAM2 cannot be directly applied to SAM3, highlighting five scientific dimensions that fundamentally separate the two models. As illustrated in the mindmap, the first major divergence arises from the prompt modalities: SAM2 relies exclusively on visual prompts such as clicks or bounding boxes, whereas SAM3 interprets natural-language descriptions and concept exemplars, requiring semantic grounding rather than spatial anchoring. The second branch emphasizes the architectural break, where SAM2 is limited to a vision-plus-temporal-memory pipeline, while SAM3 incorporates a full vision-language fusion mechanism capable of open-vocabulary reasoning. The third distinction involves training objectives; SAM2 optimizes mask quality and temporal stability, whereas SAM3 introduces contrastive, multimodal alignment, and concept-grounding losses. The mindmap also highlights dataset asymmetry: SAM2 trains on large-scale video mask datasets like SA-V, while SAM3 requires multimodal concept annotations linking text and pixel-level regions. Finally, evaluation metrics diverge sharply SAM2 is assessed through spatial and temporal consistency, while SAM3 requires semantic, attribute-based, and open-vocabulary evaluation. Together, Figure \ref{fig:sam2sam3_mindmap} demonstrates that the SAM2-to-SAM3 gap is not incremental but structural, demanding entirely different expertise for effective training and deployment.

\subsection{SAM2: A Pure Vision-Temporal Architecture}
SAM2 extends SAM1 by introducing a temporal key value memory that supports long-sequence video reasoning and maintains object identity across frames. It combines ViT-based visual tokens with a memory retrieval pathway to propagate masks over time. The learning paradigm optimizes mask consistency, spatial coherence, and temporal stability. Architecturally, SAM2 is still purely vision-based. It lacks a text encoder, has no multimodal fusion layer, and does not maintain any concept embedding pipeline or semantic grounding mechanism. There is no open-vocabulary classification head. All training heuristics, including learning rate schedules, loss shaping, prompt augmentation, and temporal smoothing, are constrained to geometric and temporal signals. This narrow focus means that SAM2 expertise primarily concerns how to design prompts, how to stabilize masks across frames, and how to manage memory capacity and latency.

\subsection{SAM3: A Unified Multimodal Vision Language System}
\begin{figure}[h!]
     \centering
     \includegraphics[width=0.95\linewidth]{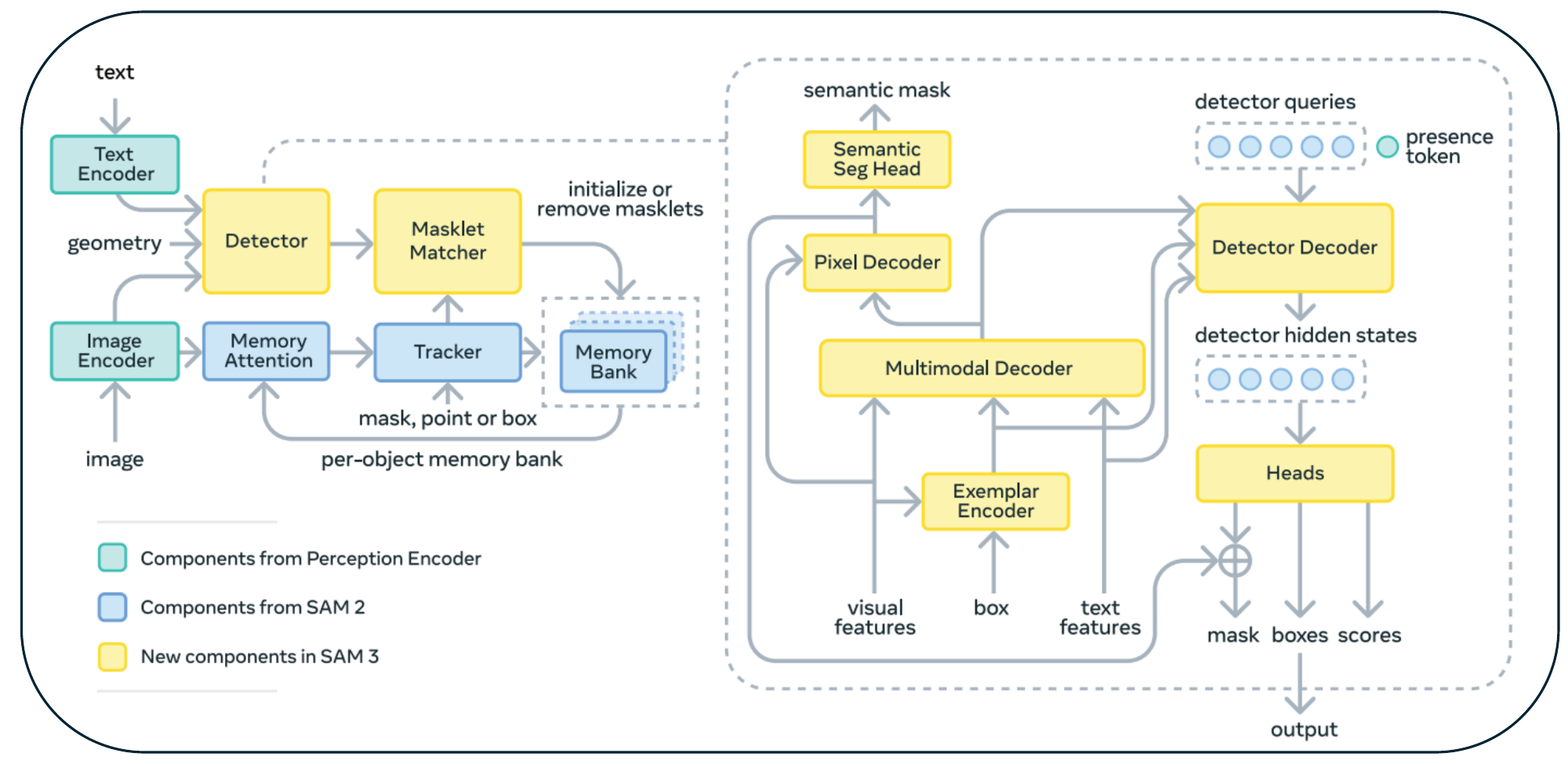}
    \caption{SAM3 architecture, highlighting new multimodal components in yellow, inherited SAM2 \cite{ravi2024sam} modules in blue, and the Perception Encoder \cite{bolya2025perception} in cyan. The model integrates vision, text, geometry, and exemplar prompts through a dual encoder–decoder transformer, enabling concept-level segmentation beyond SAM2’s purely prompt-driven capabilities. Image sourced from \cite{carion2025sam}}
    \label{fig:archSAM3}
\end{figure}

SAM3 introduces a unified multimodal architecture as illustrated in Figure \ref{fig:archSAM3} that integrates a large-scale text encoder, a semantic alignment module, and a concept-conditioned mask decoder with the vision backbone. In typical implementations, a transformer-based image encoder produces visual tokens, while a separate language model, such as a LLaMA- or Qwen-family encoder, generates text embeddings for prompts. A multimodal alignment layer fuses these two token streams through cross-attention, enabling the model to associate phrases like ``ripe apple'' or ``diseased leaf'' with specific visual patterns. Training SAM3 requires optimizing joint objectives: segmentation mask loss, cross-modal embedding loss, contrastive concept alignment loss, and semantic grounding consistency. These losses are fundamentally different from SAM2's temporal tracking objectives. Knowledge of prompt interactions, memory behavior, and click-based refinement in SAM2 offers little advantage when working with SAM3, which is dominated by the challenges of multimodal representation learning and concept-level reasoning.

SAM3 introduces a fundamentally expanded architecture that generalizes the prompt-driven segmentation paradigm of SAM2 into a fully multimodal, concept-aware vision-language system. As illustrated in Figure~\ref{fig:archSAM3}, the architecture integrates three color-coded components: new SAM3 modules highlighted in yellow, inherited SAM2 modules shown in blue, and the Perception Encoder (PE) from Bolya et al. (2025)\cite{bolya2025perception} in cyan reflecting the shift from geometric prompting to semantic concept reasoning. Whereas SAM2 relies on a single vision encoder and mask decoder, SAM3 employs a dual encoder–decoder transformer architecture that simultaneously processes vision, text, geometric prompts, and exemplar images \cite{carion2025sam}.

\paragraph{Vision--Language Encoders and PE Backbone.}  
SAM3 uses a 450M-parameter vision encoder and a 300M-parameter causal text encoder trained on 5.4B image--text pairs via contrastive learning \cite{carion2025sam}. The contrastive objective aligns visual embeddings $v$ and text embeddings $t$ in a shared semantic space:
\begin{equation}
\mathcal{L}_{\text{con}} = -\log 
\frac{\exp(\langle v, t \rangle / \tau)}
{\sum_j \exp(\langle v, t_j \rangle / \tau)}
\label{eq:contrastive}
\tag{Eq. 1}
\end{equation}
As illustrated in Figure~\ref{fig:archSAM3}, the PE backbone processes each frame once and produces unconditioned tokens for the fusion encoder, unlike SAM2 which repeatedly conditions computation on temporal memory.

\paragraph{Geometry and Exemplar Encoder.}  
SAM3 introduces a geometry exemplar encoder that generates ``geometry tokens'' (for points/boxes) and ``exemplar tokens'' (from positive or negative visual examples). These tokens attend to one another and to PE embeddings, enabling concept conditioning even without explicit manual prompts.

\paragraph{Fusion Encoder.}  
The fusion encoder forms the core multimodal module: it integrates text tokens, geometry/exemplar tokens, and frame embeddings using six transformer blocks with cross-attention. This module transforms prompt inputs into \emph{semantic conditioning signals} that guide mask generation, a capability fundamentally absent in SAM2's architecture.

\paragraph{DETR-style Decoder and Object Queries.}  
SAM3 adopts a DETR-inspired decoder with $Q = 200$ learned object queries. These queries self-attend, cross-attend to conditioned frame embeddings, and employ iterative refinement techniques for improved localization. The probability that a query matches a concept is factorized as shown in equation 2:
\begin{equation}
p(\text{query}_i \rightarrow \text{concept}) =
p(\text{local match}) \cdot p(\text{concept present in image}),
\label{eq:presence}
\tag{Eq. 2}
\end{equation}
introducing a \emph{presence head} that stabilizes semantic classification under ambiguity.

\paragraph{Ambiguity Handling with Mixture-of-Experts (MoE).}  
To address semantic ambiguity, SAM3 uses a mixture-of-experts (MoE) segmentation head with winner-takes-all (WTA) optimization (Equation 3):
\begin{equation}
L_{\text{WTA}} = L_{k^\star}, \qquad 
k^\star = \arg\min_k L_k.
\label{eq:wta}
\tag{Eq. 3}
\end{equation}
A separate classification head selects the correct expert at inference, enabling SAM3 to choose among competing interpretations of ambiguous prompts something SAM2 cannot do.

\paragraph{Why Prompt-Based Expertise Fails for Concept Segmentation.}  
SAM2 expertise does not transfer to SAM3 because SAM2 optimizes only geometric losses (Equation: 4):
\begin{equation}
\mathcal{L}_{\text{mask}} = 1 - \text{IoU}(M_{\text{pred}}, M_{\text{gt}}), 
\qquad
\mathcal{L}_{\text{temp}} = \| M_t - M_{t-1} \|_2,
\label{eq:sam2}
\tag{Eq. 4}
\end{equation}
where segmentation is guided strictly by spatial prompts.  
SAM3 instead optimizes multimodal alignment, semantic grounding, ambiguity resolution, and concept presence estimation. The operational question shifts from \emph{``Where is the object?''} to \emph{``What semantic concept does this region represent?''}.

This transition is architectural, mathematical, and cognitive in nature making SAM2 prompt expertise insufficient for SAM3's concept-driven segmentation paradigm.

\paragraph{Summary of Key Scientific Reasons for Non-Transferability :}

\begin{figure}[h!]
     \centering
     \includegraphics[width=0.95\linewidth]{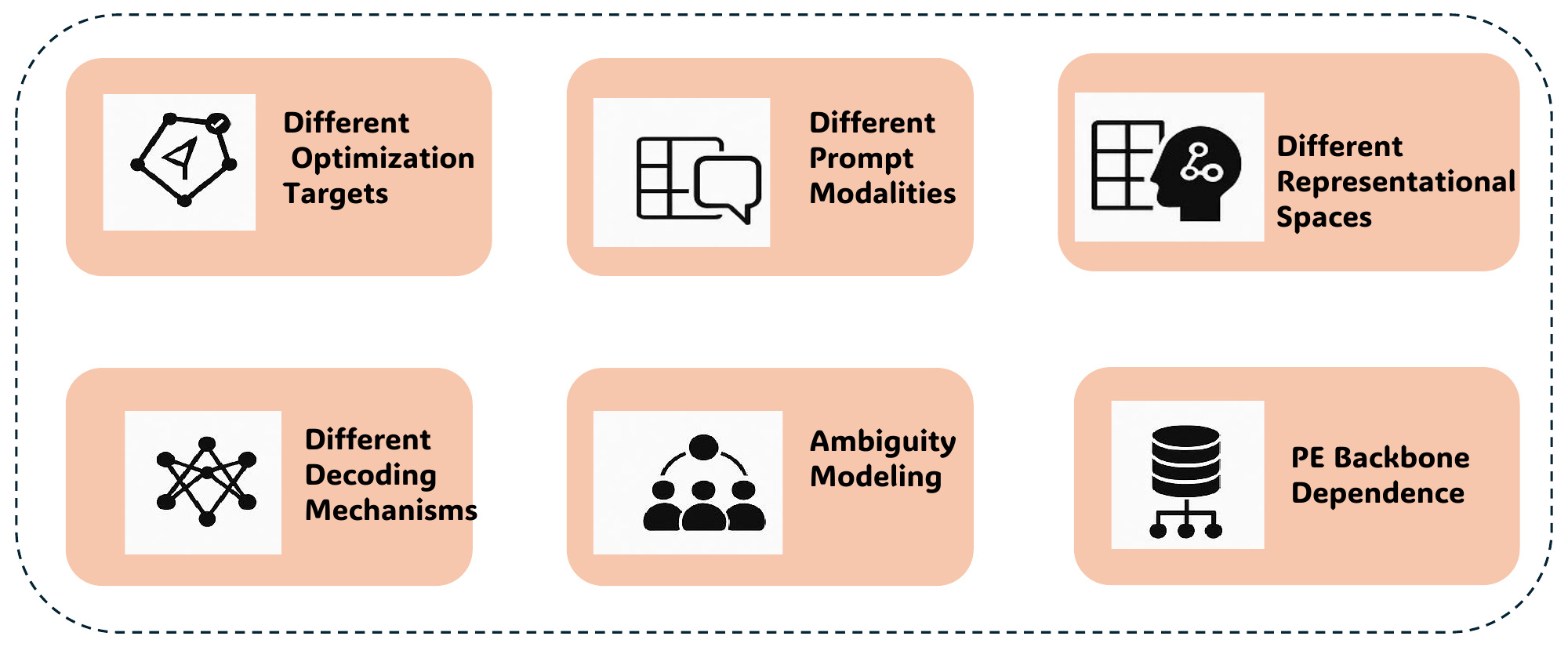}
    \caption{Scientific overview of the six core reasons SAM2 expertise fails to transfer to SAM3. The infographic highlights differences in optimization objectives, multimodal prompting, semantic embedding spaces, DETR-style decoding, ambiguity modeling, and PE-driven representation learning illustrating the fundamental architectural and conceptual discontinuity between prompt-based and concept-driven segmentation.}
    \label{fig:nontransfer} 
\end{figure}

Figure~\ref{fig:nontransfer} visually summarizes the six core scientific reasons why expertise developed for SAM2 does not transfer to SAM3. The diagram highlights fundamental differences in optimization goals, prompt modalities, representational spaces, decoding mechanisms, ambiguity handling, and reliance on PE semantic embeddings. Together, these distinctions illustrate the architectural and functional discontinuity between the two models and motivate the detailed points discussed below in six points:

\begin{itemize}
\item \textbf{Different Optimization Targets:} SAM2 optimizes geometry; SAM3 optimizes semantics and multimodal alignment.
\item \textbf{Different Prompt Modalities:} SAM2 uses points/boxes; SAM3 uses text, exemplars, and multimodal prompts.
\item \textbf{Different Representational Spaces:} SAM2 operates in pixel space; SAM3 aligns vision and language embeddings.
\item \textbf{Different Decoding Mechanisms:} SAM3’s DETR-style decoder reasons over concepts rather than object outlines.
\item \textbf{Ambiguity Modeling:} SAM3 uses MoE and presence prediction structures absent in SAM2.
\item \textbf{PE Backbone Dependence:} SAM3 relies on PE-trained embeddings that encode semantics not available to SAM2.
\end{itemize}

\section{Dataset and Annotation Differences}
\subsection{SAM2 and the SA-V Dataset}
SAM2 is trained primarily on large-scale video datasets with dense segmentation masks. The SA-V (\href{https://github.com/facebookresearch/sam2/blob/main/sav_dataset/README.md}{Source Link)} dataset is an example of such a corpus: it provides millions of video clips paired with pixel-level mask annotations. These masks encode object boundaries and shapes but do not necessarily include rich semantic or attribute-level metadata. No text labels or concept annotations are required. As a result, the model learns to optimize spatial consistency, boundary quality, and temporal propagation of objects across frames. SAM2 learns robust objectness and temporal tracking but cannot distinguish between conceptual categories such as ripe versus unripe fruit, healthy versus diseased foliage, or different defect types on a manufacturing line. Training procedures developed for SAM2 revolve around improving temporal robustness, handling occlusion, and stabilizing propagation across long sequences rather than mastering semantic distinctions.

\subsection{SAM3 and the Multimodal Concept Dataset Requirement}
SAM3, in contrast to SAM2’s purely geometric training paradigm, depends on datasets that explicitly bind textual concepts to visual scenes and pixel-level masks, forming the foundation for its multimodal reasoning capabilities. As summarized in Table \ref{tab:sam_dataset_comparison}, SAM3’s training corpora must provide not only segmentation masks but also rich linguistic descriptions and concept labels that capture object categories, attributes, states, and contextual semantics. These labels include fine-grained descriptors such as “ripe apple,” “healthy leaf,” “stem,” “bruised fruit,” or “immature blossom,” enabling the model to associate nuanced textual tokens with corresponding visual patterns. Because SAM3 learns through cross-modal alignment mapping text embeddings to vision embeddings the dataset must reflect the conceptual variety and semantic ambiguity present in real-world environments. This requirement introduces substantial challenges absent in SAM2. For example, concept ambiguity becomes a central difficulty when terms such as “ripe,” “healthy,” or “damaged” vary by domain-specific definitions, measurement standards, or cultural interpretations. A “ripe apple” in one cultivar, orchard system, or lighting condition may differ significantly in visual appearance from another, requiring large, diverse annotations to avoid confounding the model’s semantic grounding. 

Furthermore, as Table \ref{tab:sam_dataset_comparison} illustrates, the multimodal datasets used for SAM3 include millions of noun-phrase annotations, hard-negative examples, and cross-domain concept variations spanning agriculture, medical imaging, industrial inspection, robotics, and biological sciences. This diversity is essential to prevent overfitting to narrow visual-linguistic correlations, yet it complicates dataset construction and quality control. Multilingual expressions, synonym sets, polysemous terms, and subtle linguistic nuances further expand the semantic space the model must master. Additionally, SAM3 must learn attribute-level distinctions such as color, texture, material state, or morphological condition that often require human expertise or context-specific annotation protocols. These complexities transform SAM3 training into a multimodal alignment and semantic grounding problem rather than a purely geometric segmentation task. The model must learn not only where objects are located but what they represent, how they differ, and how textual descriptions relate to those differences. As a result, SAM3’s training demands significantly more expressive datasets, sophisticated annotation pipelines, and robust concept validation mechanisms, making its data requirements far more intricate and scientifically demanding than those of SAM2.

\begin{table*}[t]
\centering
\caption{Comprehensive comparison of datasets and annotation pipelines for SAM2\cite{ravi2024sam} and SAM3\cite{carion2025sam}. SAM2 relies on video mask propagation datasets such as SA-V (\href{https://github.com/facebookresearch/sam2/blob/main/sav_dataset/README.md}{Source Link)}, whereas SAM3 requires large-scale multimodal concept datasets with text–mask alignment, as documented in the SA-Co dataset engine (\href{https://docs.ultralytics.com/models/sam-3/}{Source Link}).}
\begin{tabular}{p{3.4cm} p{5.4cm} p{5.4cm}}
\toprule
\textbf{Category} & \textbf{SAM2 (SA-V Dataset)} & \textbf{SAM3 (SA-Co Dataset Family)} \\
\midrule
\textbf{Primary Data Type} & Videos with dense pixel masks; no text labels & Images + videos paired with noun-phrase concepts, masks, and hard negatives \\
\midrule
\textbf{Scale (Images)} & Not explicitly reported; focus on video masklets & 5.2M HQ images with 4M unique noun phrases (SA-Co/HQ) \\
\midrule
\textbf{Scale (Videos)} & SA-V: millions of video clips with dense masks & 52.5K annotated videos, 24.8K unique concepts (SA-Co/VIDEO) \\
\midrule
\textbf{Total Masks} & Millions of frame-level masks propagated temporally & 52M high-quality human+AI-verified masks; 1.4B synthetic masks \\
\midrule
\textbf{Concept Labels} & None (pure geometric masks) & 4M unique NPs; 207K benchmark concepts; includes hard negatives \\
\midrule
\textbf{Annotation Source} & Human-generated masks + SAM2 propagation & Human annotators + AI annotators (LLMs + SAM3 proposals) \\
\midrule
\textbf{Annotation Pipeline} & Video masklet propagation; temporal refinement & 4-phase data engine: NP proposal → mask proposal → mask verification → exhaustivity verification → correction \\
\midrule
\textbf{Semantic Metadata} & Not included; no attributes, states, or categories & Includes attributes (e.g., ``ripe apple''), states, objects, fine-grained categories \\
\midrule
\textbf{Ambiguity Handling} & None; geometry-only & Multi-annotator labels, ambiguity module, ontology-based NP curation \\
\midrule
\textbf{Modeling Objective Enabled} & Temporal objectness, mask propagation & Open-vocabulary concept segmentation; text–vision alignment \\
\midrule
\textbf{Benchmark Coverage} & SA-V videos & SA-Co/Gold, Silver, Bronze, Bio, VEval benchmarks \\
\midrule
\textbf{Example Dataset Stats} & No explicit concept count; mask-only & 207K unique phrases, 121K media items, 3M media–phrase pairs \\
\midrule
\textbf{Data Diversity} & Primarily natural videos with object motion & 15+ visual domains; curated ontology with 22.4M nodes \\
\bottomrule
\end{tabular}
\label{tab:sam_dataset_comparison}
\end{table*}

\section{Training and Hyperparameter Distinctions}
\subsection{Why SAM2 Training Knowledge Does Not Apply to SAM3}
SAM2 hyperparameters are designed to optimize a vision-temporal pipeline. Typical tuning targets include the strength and depth of temporal memory, the configuration of key value embeddings, the size of attention windows, frame sampling strategies, and the resolution of visual prompts and masks \cite{ke2023segment, wang2023samrs, ding2025sam2long}. Loss balancing in SAM2 focuses on trade-offs between segmentation accuracy and temporal stability \cite{sun2025efficient, fan2025research}, with additional attention to preventing drift or fragmentation over long video sequences \cite{yin2025improvement, liu2025resurgsam2}. These design patterns assume that prompts are visual, supervision is geometric, and evaluation is based on pixel overlap or identity tracking.

SAM3 introduces a different class of hyperparameters. Trainers must set learning rates for the text encoder and image encoder, adjust the relative weighting of segmentation loss versus cross-modal contrastive loss, tune the temperature parameters in contrastive objectives, and determine the depth at which visual and textual tokens are fused. The model also depends on the scale assigned to semantic grounding losses that penalize misalignment between concepts and visual evidence. Even the shape of optimization schedules changes: freezing or partially freezing the text encoder during early training, staged unfreezing, and curriculum strategies on concept complexity become relevant. None of these design decisions are encountered in SAM2 training. As a result, experience in SAM2 fine-tuning, which focuses on temporal memory and spatial prompts, does not translate directly into the multimodal, concept-alignment-centric world of SAM3.

\subsection{Divergent Data Augmentation and Optimization Strategies}
The divergence extends to data augmentation and optimization strategies. SAM2 benefits mainly from geometric augmentations such as random scaling, cropping, flipping, and affine transformations that improve robustness to motion, viewpoint changes, and occlusion \cite{ding2025pvuw, zhao2024inspiring, jiaxing2025sam2}. Color jitter and brightness changes are generally helpful but primarily in the context of preserving temporal consistency \cite{wang2024empirical}. For SAM2, the key objective is to maintain stable masks and object identities across diverse video conditions \cite{zhou2025sam2}.

\begin{figure}[ht]
\centering
\begin{tikzpicture}[
    node distance=8mm and 20mm,
    coltitle/.style={font=\sffamily\bfseries\small},
    sideboxSAM2/.style={
        rectangle, rounded corners, draw,
        align=left, font=\footnotesize,
        minimum width=5.0cm, minimum height=1.1cm,
        inner sep=2.5pt, fill=red!20
    },
    sideboxSAM3/.style={
        rectangle, rounded corners, draw,
        align=left, font=\footnotesize,
        minimum width=5.0cm, minimum height=1.1cm,
        inner sep=2.5pt, fill=green!30
    },
    arrowstyle/.style={-{Stealth}, thick}
]

% Column titles
\node[coltitle] (sam2title) {SAM2: Geometric \& Temporal-Stability};
\node[coltitle, right=40mm of sam2title] (sam3title)
    {SAM3: Semantic \& Multimodal-Stability};

% Row 1: Augmentation philosophy
\node[sideboxSAM2, below=7mm of sam2title] (s2r1) {1) Geometric augmentations:\\
-- random scaling, cropping, flipping, affine transforms\\
-- robustness to motion, viewpoint, occlusion};
\node[sideboxSAM3, right=20mm of s2r1] (s3r1) {1) Semantic-preserving augmentations:\\
-- careful color/structure changes\\
-- avoid distorting attributes (e.g., ripeness, damage)};

\draw[arrowstyle] (s2r1.east) -- (s3r1.west);

% Row 2: Photometric / context handling
\node[sideboxSAM2, below=8mm of s2r1] (s2r2) {2) Mild photometric changes:\\
-- color jitter / brightness for temporal consistency\\
-- primarily to stabilize video masks};
\node[sideboxSAM3, right=20mm of s2r2] (s3r2) {2) Context-aware cropping \& color:\\
-- must preserve global scene context (tree canopy, branches)\\
-- maintain text--image semantic alignment};

\draw[arrowstyle] (s2r2.east) -- (s3r2.west);

% Row 3: Augmentation objective
\node[sideboxSAM2, below=8mm of s2r2] (s2r3) {3) Augmentation objective:\\
-- maintain stable masks and IDs across frames\\
-- optimize temporal consistency of object tracks};
\node[sideboxSAM3, right=20mm of s2r3] (s3r3) {3) Augmentation objective:\\
-- preserve concept meaning and attribute semantics\\
-- support open-vocabulary, concept-level segmentation};

\draw[arrowstyle] (s2r3.east) -- (s3r3.west);

% Row 4: Optimization complexity
\node[sideboxSAM2, below=12mm of s2r3] (s2r4) {4) Optimization strategy:\\
-- standard mixed-precision training\\
-- moderate memory footprint from vision-only backbone};
\node[sideboxSAM3, right=20mm of s2r4] (s3r4) {4) Optimization strategy:\\
-- mixed-precision + gradient checkpointing\\
-- high memory footprint from joint vision--language processing};

\draw[arrowstyle] (s2r4.east) -- (s3r4.west);

% Row 5: Batching and scheduling
\node[sideboxSAM2, below=8mm of s2r4] (s2r5) {5) Batching/scheduling:\\
-- simpler batching over video clips\\
-- schedules tuned for temporal mask losses};
\node[sideboxSAM3, right=20mm of s2r5] (s3r5) {5) Batching/scheduling:\\
-- careful batching over image--text pairs and concepts\\
-- schedules balance segmentation and contrastive losses};

\draw[arrowstyle] (s2r5.east) -- (s3r5.west);

% Row 6: Hyperparameter search
\node[sideboxSAM2, below=8mm of s2r5] (s2r6) {6) Hyperparameter search:\\
-- focused on temporal stability and IoU\\
-- relatively low-dimensional search space};
\node[sideboxSAM3, right=20mm of s2r6] (s3r6) {6) Hyperparameter search:\\
-- involves fusion depth, loss weights, text LR, temperature\\
-- no direct analogue in SAM2 pipelines};

\draw[arrowstyle] (s2r6.east) -- (s3r6.west);

\end{tikzpicture}
\caption{Divergent data augmentation and optimization strategies for SAM2\cite{ravi2024sam} and SAM3\cite{carion2025sam}. SAM2 emphasizes geometric and temporal robustness for video masks, whereas SAM3 requires semantic-preserving augmentations and multimodal optimization tuned for concept-level, open-vocabulary segmentation.}
\label{fig:sam2_sam3_augment_opt}
\end{figure}
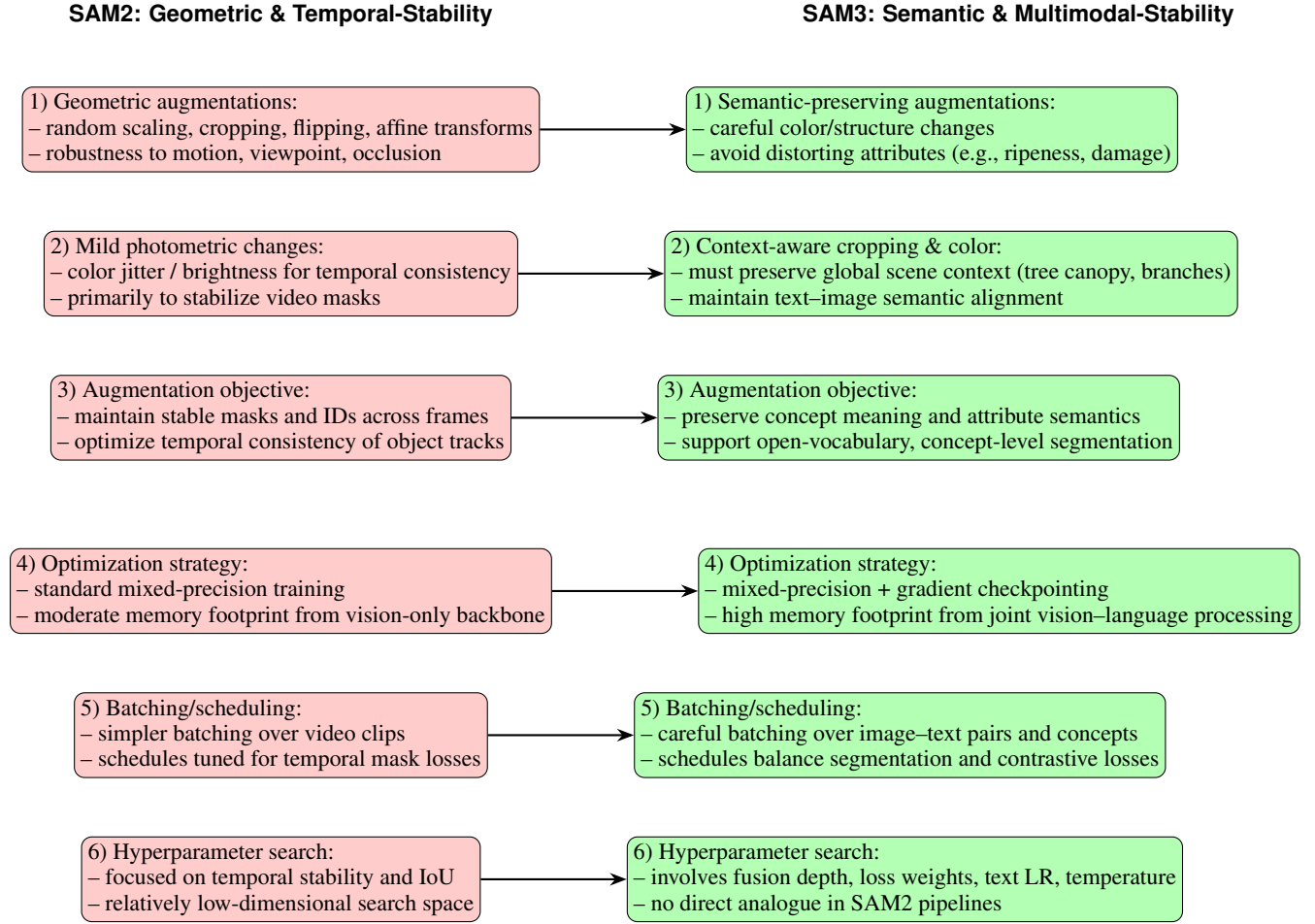

In contrast, SAM3 requires data augmentation strategies that preserve semantic meaning while still improving robustness, creating a fundamentally different design landscape from SAM2. As illustrated in Figure~\ref{fig:sam2_sam3_augment_opt}, transformations that were harmless for SAM2—such as strong color jitter, aggressive cropping, or heavy affine distortions can degrade SAM3’s ability to interpret attributes like ripeness, damage severity, material type, or structural context. Because SAM3 aligns visual features with linguistic concepts, any augmentation that disrupts color fidelity, texture cues, or global spatial context risks breaking the link between text prompts and visual semantics. This makes concept-preserving augmentation essential for stable training. 

Beyond augmentation, SAM3 also introduces new optimization challenges. Multimodal fusion substantially increases memory and compute requirements, making mixed-precision training, gradient checkpointing, and careful micro-batching necessary for scalable fine-tuning. Hyperparameters now include contrastive-loss temperatures, cross-modal fusion depth, and segmentation–alignment loss weights—components absent in SAM2’s simpler geometric optimization. As shown again in Figure~\ref{fig:sam2_sam3_augment_opt}, these multimodal constraints make SAM3 hyperparameter search more complex, non-linear, and highly sensitive to semantic drift. Overall, the augmentation and optimization divergence reinforces the broader architectural discontinuity between prompt-driven SAM2 and concept-driven SAM3.

\section{Evaluation, Metrics, and Failure Modes}
\subsection{SAM2 Evaluation Metrics}
SAM2 is evaluated using metrics that focus on geometric segmentation and temporal performance. Common metrics include intersection over union (IoU) for mask overlap, boundary precision and recall for edge quality, and temporal stability indices that measure the consistency of masks frame-to-frame. In video segmentation benchmarks, identity preservation metrics assess whether the same object is tracked correctly across time. These metrics are well suited to scenarios such as video editing, object tracking, and interactive segmentation, where the user has already specified which object to follow. However, they do not measure whether the model has correctly understood a semantic concept; they simply measure how accurately and consistently the model follows a prompted region.

\subsection{SAM3 Evaluation Metrics}
SAM3 introduces additional evaluation dimensions that focus on semantic correctness and concept-level understanding. The model must be judged not only by pixel-level IoU but also by how well it recalls all instances of a concept given a text prompt. Concept recall measures the proportion of relevant objects that are successfully segmented. Semantic localization error quantifies how much of the predicted mask corresponds to the intended concept versus irrelevant regions. Open-vocabulary generalization metrics evaluate performance on concepts that were not present in the training set or are used in new combinations. Attribute segmentation accuracy is important for prompts involving properties such as ``ripe'', ``damaged'', or ``healthy''. Sensitivity to language ambiguity, synonym choices, and phrasing must also be considered. These metrics have no counterpart in SAM2 evaluation pipelines and require new benchmark design. Consequently, evaluation expertise built around SAM2 cannot be directly reused for SAM3 without incorporating these semantic and multimodal dimensions.

\section{Comprehensive Tables Describing the SAM2-to-SAM3 Gap}
\begin{table}[h!]
\centering
\caption{Architectural and training differences between SAM2\cite{ravi2024sam} and SAM3\cite{carion2025sam}.}
\begin{tabular}{p{2.8cm} p{6.2cm} p{7.7cm}}
\toprule
\textbf{Category} & \textbf{SAM2} & \textbf{SAM3} \\
\midrule
Primary Modality & Vision-only (video) & Vision + text + example images \\
Prompt Type & Points, boxes, masks & Natural language prompts and concept exemplars \\
Reasoning Type & Geometric and temporal & Semantic, conceptual, and multimodal \\
Architecture & ViT backbone + temporal memory & ViT backbone + text encoder + fusion module \\
Training Losses & Segmentation and temporal tracking losses & Segmentation, contrastive, semantic grounding losses \\
Dataset Type & Video with pixel masks & Multimodal concept-annotated datasets \\
Optimization Focus & Memory stability and propagation & Cross-modal alignment and concept fidelity \\
\bottomrule
\end{tabular}
\end{table}

\begin{table}[h!]
\centering
\caption{Evaluation differences between SAM2\cite{ravi2024sam} and SAM3\cite{carion2025sam}.}
\begin{tabular}{p{6.4cm} p{3.2cm} p{2.2cm}}
\toprule
\textbf{Metric Type} & \textbf{SAM2} & \textbf{SAM3} \\
\midrule
Mask IoU & Yes & Yes \\
Temporal Tracking Metrics & Yes & No \\
Concept Recall & No & Yes \\
Semantic Grounding Error & No & Yes \\
Open-Vocabulary Generalization & No & Yes \\
Attribute Segmentation Accuracy & No & Yes \\
Prompt Sensitivity (Visual) & High & Medium \\
Language Ambiguity Sensitivity & Not applicable & High \\
\bottomrule
\end{tabular}
\end{table}

Table \ref{tab:equations_sam2_sam3} provides a consolidated scientific comparison of SAM2 and SAM3, highlighting the mathematical, architectural, and evaluative differences that define the SAM2-to-SAM3 transition. As shown in the table, SAM2 is governed primarily by geometric losses such as IoU and temporal stability, whereas SAM3 introduces contrastive alignment, semantic grounding, and concept-conditioned segmentation objectives. These distinctions are not merely architectural but fundamentally reshape how segmentation is optimized, supervised, and interpreted. The evaluation metrics listed in Table \ref{tab:equations_sam2_sam3} further demonstrate why SAM3 requires entirely new assessment protocols centered on semantic fidelity and open-vocabulary reasoning. Collectively, this table underscores why expertise developed for SAM2 cannot be directly transferred to multimodal, concept-driven SAM3 workflows.

\begin{table*}[ht!]
\centering
\caption{Scientific comparison of SAM2\cite{ravi2024sam} and SAM3\cite{carion2025sam}: core equations, training objectives, and evaluation metrics. Boundary lines separate each conceptual block for clarity.}
\scriptsize 
\begin{tabular}{p{1.7cm} p{6.8cm} p{6.8cm}}
\toprule
\textbf{Category} & \textbf{SAM2 (Prompt-Based Vision Model)} & \textbf{SAM3 (Concept-Driven Multimodal Model)} \\
\midrule

\textbf{Primary Objective} 
& Spatial mask segmentation and temporal propagation 
& Concept-level segmentation conditioned on language, attributes, or exemplars \\
\midrule

\textbf{Core Loss Function} 
& \begin{tabular}[t]{@{}l@{}}
Mask Loss: $\mathcal{L}_{\text{mask}} = 1 - \text{IoU}(M_{\text{pred}}, M_{\text{gt}})$ \\
Temporal Consistency: $\mathcal{L}_{\text{temp}} = \|M_t - M_{t-1}\|_2$
\end{tabular}
& \begin{tabular}[t]{@{}l@{}}
Segmentation Loss: $\mathcal{L}_{\text{seg}} = 1 - \text{IoU}(M_{\text{pred}}, M_{\text{concept}})$ \\
Contrastive Alignment: \\
$\mathcal{L}_{\text{con}} = -\log \frac{\exp(\langle v, t \rangle / \tau)}{\sum_j \exp(\langle v, t_j \rangle / \tau)}$ \\
Grounding Loss: $\mathcal{L}_{\text{ground}} = \|f_{\text{vision}}(x)-f_{\text{text}}(c)\|_2$
\end{tabular} \\
\midrule

\textbf{Optimization Target} 
& Reduce geometric error; stabilize temporal mask propagation 
& Maximize semantic alignment; improve concept fidelity \\
\midrule

\textbf{Prompt Input Type}
& Visual prompts (points, boxes, masks) 
& Text prompts, concept exemplars, image–text fusion \\
\midrule

\textbf{Architecture}
& ViT Backbone + Temporal Memory 
& ViT Backbone + Text Encoder + Cross-Modal Fusion + Concept Decoder \\
\midrule

\textbf{Dataset  Requirements}
& Video datasets with dense pixel masks (SA-V) 
& Multimodal datasets linking segmentation masks with textual concept labels \\
\midrule

\textbf{Concept Embeddings}
& No semantic embedding: $f_{\text{text}}$ absent 
& Joint embeddings: $f_{\text{vision}}(x)$ aligned with $f_{\text{text}}(c)$ \\
\midrule

\textbf{Key Evaluation Metrics}
& \begin{tabular}[t]{@{}l@{}}
Mask IoU: $\text{IoU} = \frac{|M_p \cap M_g|}{|M_p \cup M_g|}$ \\
Boundary F1 Score \\
Temporal Identity Preservation
\end{tabular}
& \begin{tabular}[t]{@{}l@{}}
Concept Recall: $\frac{TP}{TP+FN}$ \\
Semantic Grounding Error: $\|B_{\text{pred}} - B_{\text{concept}}\|_1$ \\
Open-Vocabulary F1; Attribute Accuracy
\end{tabular} \\
\midrule

\textbf{Failure Modes}
& Prompt sensitivity; memory drift; occlusion-induced degradation 
& Embedding misalignment; language ambiguity; semantic leakage \\
\bottomrule
\end{tabular}
\label{tab:equations_sam2_sam3}
\end{table*}

\section{Conclusion and Future Directions: Toward a Concept-Driven Segmentation Era}
The transition from SAM2 to SAM3 represents a foundational redefinition of what it means to “segment” in modern computer vision. Rather than an incremental improvement, SAM3 introduces a paradigm shift from geometric prompting to semantic reasoning, fundamentally altering how segmentation models understand visual content. SAM2 users skilled in designing visual prompts, stabilizing temporal memory, and optimizing mask propagation possess expertise that is deeply rooted in spatial interactions. In contrast, SAM3 demands proficiency in multimodal alignment, linguistic embeddings, semantic grounding, concept disambiguation, and large-scale vision-language pretraining. This shift moves segmentation from answering “Where is the object?” to addressing “What concept does this region represent?”, which disrupts long-standing assumptions about prompt design, dataset construction, and evaluation methodology.

\begin{figure}[ht!]
\centering
\begin{tikzpicture}[
    mindmap,
    grow cyclic,
    concept color=blue!35,
    every node/.style={concept, circular drop shadow, font=\small \bfseries, text=black},
    level 1 concept/.append style={level distance=4cm, sibling angle=72}
]

\node{Toward a\\Concept-Driven\\Segmentation Era}
    child[concept color=green!40] {
        node {From Spatial\\Prompts to\\Concept Prompts}
    }
    child[concept color=orange!40] {
        node {Multimodal\\Data:\\Text + Masks}
    }
    child[concept color=purple!40] {
        node {New Training\\Objectives:\\Contrastive \&\\Grounding}
    }
    child[concept color=red!40] {
        node {Semantic\\Evaluation:\\Concept Recall,\\Open-Vocab F1}
    }
    child[concept color=cyan!40] {
        node {Impact on\\Applications:\\Robotics,\\Medical, Agriculture}
    };

\end{tikzpicture}
\caption{Compact mindmap summarizing the shift toward a concept-driven segmentation era, emphasizing the move from spatial prompts to concept prompts, multimodal data, new training objectives, semantic evaluation, and downstream application impact.}
\label{fig:concept_driven_segmentation_mindmap}
\end{figure}
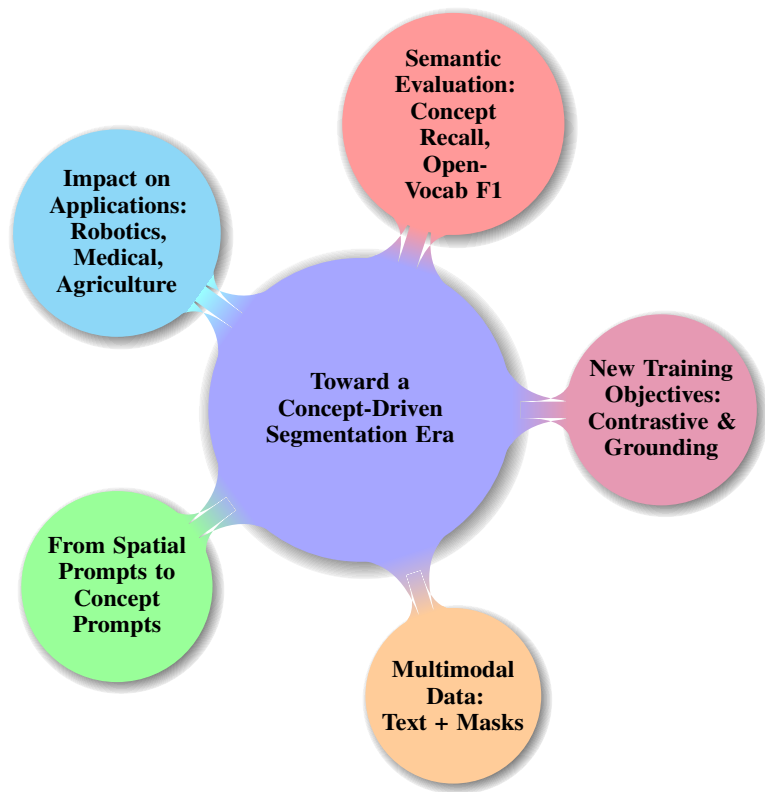

SAM3 requires researchers to build and curate multimodal datasets where each mask is linked not only to object identity but to concept-level attributes, synonyms, and contextual meanings. Its training pipeline optimizes for semantic alignment losses, presence modeling, ambiguity resolution, and cross-attention fusion loss terms and architectural components that have no analog in SAM2. In effect, SAM3 changes both the computational substrate and the cognitive framing of segmentation. This explains why prompt-based workflows from SAM2 cannot be transferred: the skill set, objectives, and representational assumptions are fundamentally different. Where SAM2 teaches a model how to follow, SAM3 teaches a model how to understand.

Looking ahead, this conceptual leap will reshape research and industry practices across robotics, agriculture, medical imaging, manufacturing, autonomous systems, and multimodal AI. The broader community adopting SAM3 must prepare for three major transformations. First, dataset engineering will evolve toward richer semantic annotations, possibly requiring new tools for scalable concept labeling and cross-domain knowledge transfer. Second, evaluation protocols will increasingly emphasize open-vocabulary generalization, linguistic robustness, and semantic grounding in addition to traditional IoU metrics. Third, successful deployment of SAM3 in real-world systems will require interdisciplinary expertise integrating computer vision, natural language processing, and multimodal representation learning. Future research should explore more efficient training strategies for multimodal segmentation, including lightweight adapters, domain-specific concept banks, and curriculum learning for ambiguous or hierarchical concepts. Moreover, as SAM3 becomes integrated into embodied agents and decision-making systems, ensuring transparency, interpretability, and robustness under linguistic variability will be essential. The next generation of segmentation technologies will likely move further toward unified perception-language-reasoning frameworks, where models understand not only objects and concepts but also intent and context. 

Ultimately, SAM3 marks the beginning of the semantic era in segmentation. By embracing multimodal reasoning, future researchers and practitioners can build systems that perceive the world not as regions of pixels, but as structured semantic environments paving the way for more intelligent, adaptable, and context-aware vision systems.
\section*{Declaration of Competing Interest}
The authors declare that they have no known competing financial interests or personal relationships that could have appeared to influence the work reported in this paper.

\bibliographystyle{unsrt}  
\bibliography{references}
  %%% Uncomment this line and comment out the ``thebibliography'' section below to use the external .bib file (using bibtex) .

%%% Uncomment this section and comment out the \bibliography{references} line above to use inline references.
% \begin{thebibliography}{1}

% 	\bibitem{kour2014real}
% 	George Kour and Raid Saabne.
% 	\newblock Real-time segmentation of on-line handwritten arabic script.
% 	\newblock In {\em Frontiers in Handwriting Recognition (ICFHR), 2014 14th
% 			International Conference on}, pages 417 422. IEEE, 2014.

% 	\bibitem{kour2014fast}
% 	George Kour and Raid Saabne.
% 	\newblock Fast classification of handwritten on-line arabic characters.
% 	\newblock In {\em Soft Computing and Pattern Recognition (SoCPaR), 2014 6th
% 			International Conference of}, pages 312 318. IEEE, 2014.

% 	\bibitem{hadash2018estimate}
% 	Guy Hadash, Einat Kermany, Boaz Carmeli, Ofer Lavi, George Kour, and Alon
% 	Jacovi.
% 	\newblock Estimate and replace: A novel approach to integrating deep neural
% 	networks with existing applications.
% 	\newblock {\em arXiv preprint arXiv:1804.09028}, 2018.

% \end{thebibliography}

\end{document}